\documentclass[11pt]{article}

\usepackage[preprint]{acl}

\usepackage{times}
\usepackage{latexsym}
\usepackage[T1]{fontenc}
\usepackage{iftex}
\ifPDFTeX
  \usepackage[utf8]{inputenc}
\fi
\usepackage{microtype}
\usepackage{inconsolata}

\usepackage{graphicx}
\usepackage{booktabs}
\usepackage{tabularx}
\usepackage{amsmath}
\usepackage{url}
\usepackage{multirow}
\usepackage{array}

\title{A Comparative Evaluation of BERT-Based and LLM-Based Models for Indic Sentence Similarity}

\author{
Aishwarya Gunjal\textsuperscript{1,2}\quad
Kanchan Khemnar\textsuperscript{1,2}\quad
Pawan Poojary\textsuperscript{1,2}\\
\textbf{Pranav Kulkarni}\textsuperscript{1,2}\quad
\textbf{Raviraj Joshi}\textsuperscript{2}\\
\textsuperscript{1}Pune Institute of Computer Technology, Pune, India\\
\textsuperscript{2}L3Cube Labs, Pune, India\\
\small\texttt{\{gunjalaishwarya625, kanchankhemnar170, pawanpoojary989,}\\
\small\texttt{pranavkulkarni1115, ravirajoshi\}@gmail.com}
}

\begin{document}
\maketitle
\begin{abstract}
The relative effectiveness of BERT-based encoders and large language model (LLM)-based embeddings for Indic semantic textual similarity remains insufficiently characterized. We compare eight BERT-based and five LLM-based models on the 1,500-pair STS Benchmark development set translated into ten Indic languages and on the native Marathi MahaSTS test set. IndicSBERT achieves the highest Pearson correlation in all ten translated language sets, with a macro-average of 0.843, followed by BGE-M3 at 0.792. Harrier-270M is the strongest LLM-based model on translated STS-B, with a macro-average of 0.751. On MahaSTS, IndicSBERT and Bhasha-Embed-v0 achieve Pearson correlations of 0.901 and 0.900, respectively, while EmbeddingGemma leads the LLM-based models at 0.871. The highest-scoring BERT-based model thus exceeds the highest-scoring LLM-based model in both settings, although the magnitude of the difference varies across benchmarks. These findings indicate that LLM-derived embeddings do not uniformly improve Indic sentence similarity over established encoders and motivate evaluation that accounts for language coverage, task supervision, and benchmark construction.

\end{abstract}

\section{Introduction}
Semantic textual similarity (STS) quantifies the degree of semantic equivalence between two sentences \citep{agirre2012semeval,cer2017semeval}. Sentence embedding models support efficient similarity estimation by encoding each sentence independently and comparing the resulting vectors, typically using cosine similarity \citep{reimers2019sentence}. The quality of these representations depends on their ability to preserve graded semantic distinctions across languages and domains.

BERT-based encoders and LLM-based embedding models represent two prominent approaches to this problem. BERT-based approaches obtain sentence representations through contextual encoding and pooling, often with additional sentence-level supervision; LLM-based approaches adapt foundation language models for embedding generation. Multilingual pretraining provides broad language coverage, but coverage alone does not establish suitability for fine-grained STS. This distinction is particularly relevant to Indic languages, for which the availability of training data and evaluation resources remains uneven \citep{joshi2020state}.

The emergence of LLM-based embeddings raises an empirical question: do these models offer a consistent advantage over BERT-based encoders for Indic sentence similarity? Addressing this question requires examining both variation between model families and differences within each family. Indic-specific adaptation, multilingual alignment, training objectives, and inference procedures may influence performance independently of model lineage or parameter count.

This study extends the parent project's comparison of BERT-based models and LLMs for Marathi named entity recognition \citep{ingle2026bert} to graded sentence similarity across ten Indic languages. We evaluate eight BERT-based encoders and five LLM-based embedding models in Hindi, Marathi, Bengali, Gujarati, Kannada, Malayalam, Punjabi, Tamil, Telugu, and Urdu. The evaluation combines translated STS-B, which retains the same source pairs across languages, with MahaSTS, which provides human-annotated native Marathi pairs \citep{mirashi2025mahasts}. These complementary benchmarks enable analysis of language-level performance and cross-dataset variation.

We investigate three research questions: \textbf{RQ1}, how do BERT-based and LLM-based models compare on translated Indic STS; \textbf{RQ2}, how does performance vary with Indic specialization among encoders and model size among LLM-based systems; and \textbf{RQ3}, do the observed relative strengths persist on native Marathi data? The principal contributions are:
\begin{itemize}
\item A comparative evaluation of thirteen models from two embedding families on translated STS-B in ten Indic languages and on native Marathi MahaSTS.
\item A language-level analysis of Indic-focused encoders, multilingual encoders, and LLM-based embeddings, including within-family comparisons.
\item An analysis of cross-dataset changes in model rankings and performance gaps, distinguishing empirical observations from architectural and translation-related explanations.
\end{itemize}

\section{Related Work}
\paragraph{BERT-based sentence representations.}
Transformer architectures and contextual pretraining underpin modern sentence representation learning \citep{vaswani2017attention,devlin2019bert}. Sentence-BERT introduces siamese training to produce sentence vectors that can be compared independently using cosine similarity \citep{reimers2019sentence}. SimCSE further demonstrates the value of contrastive sentence-level training, using dropout-based positive pairs in its unsupervised formulation and NLI pairs in its supervised formulation \citep{gao2021simcse}. Multilingual approaches differ in their alignment mechanisms and training objectives. XLM-R learns cross-lingual contextual representations through multilingual pretraining \citep{conneau2020unsupervised}, whereas LaBSE explicitly optimizes language-agnostic sentence embeddings \citep{feng2020language}. Multilingual knowledge distillation aligns translated sentences with embeddings from a source-language teacher \citep{reimers2020multilingual}. MPNet combines masked and permuted language-model pretraining \citep{song2020mpnet}; its multilingual sentence-embedding variants additionally rely on sentence-level adaptation. Multilingual e5 and BGE-M3 further develop contrastive embedding learning for multilingual applications \citep{wang2024e5,chen2024bge}. These differences motivate direct evaluation of sentence-level semantic discrimination rather than inference from language coverage alone.

\paragraph{Indic representations and benchmarks.}
IndicNLPSuite and MuRIL provide corpora, benchmarks, and pretrained representations for Indian languages \citep{kakwani2020indicnlpsuite,khanuja2021muril}. IndicSBERT adapts multilingual BERT using Indic natural language inference (NLI) and STS supervision \citep{deode2023l3cube}. Bhasha-Embed-v0 uses distillation and triplet training to target Hindi, English, and Romanized Hindi \citep{sukhlecha2024bhasha}; evaluation in other languages therefore assesses transfer beyond its stated scope. For evaluation, MahaSTS supplies human-annotated Marathi similarity pairs \citep{mirashi2025mahasts}, while IndicHeadline-ID assesses semantic discrimination through article--headline matching \citep{tanksale2025headline}. These resources address complementary aspects of semantic representation, with MahaSTS directly supporting graded STS evaluation.

\paragraph{LLM-based embeddings and comparative evaluation.}
LLM2Vec shows how decoder-only language models can be adapted into text encoders by enabling bidirectional attention and applying masked next-token prediction and contrastive learning \citep{behnamghader2024llm2vec}. Qwen3-Embedding adapts foundation language models through embedding-specific training \citep{zhang2025qwen3}. EmbeddingGemma and Harrier similarly provide dedicated multilingual text embeddings \citep{google2025embeddinggemma,microsoft2026harrier}. Such systems incorporate embedding supervision and should be distinguished from unadapted generative LLMs. In related work on Marathi NER, \citet{ingle2026bert} compare fine-tuned BERT-based models with general-purpose LLMs. Our study examines the same broader model-selection question for sentence similarity, using embedding-trained LLM derivatives and evaluating both translated multilingual data and native Marathi data. Accordingly, the comparison concerns complete pretrained systems rather than an isolated architectural variable.

\paragraph{Evaluation breadth and translated benchmarks.}
MTEB evaluates embeddings across multiple tasks and languages and finds that no single method dominates every task \citep{muennighoff2023mteb}. This motivates distinguishing graded STS performance from broader embedding utility. Benchmark construction also matters: \citet{artetxe2020translation} demonstrate that translation can introduce artifacts affecting cross-lingual transfer, including changes to lexical overlap in sentence-pair tasks. Their findings concern transfer settings rather than the STS comparisons reported here, but motivate complementing translated evaluation with native-language judgments. Our study focuses on this narrower question of Indic similarity across translated and native resources.

\section{Experimental Setup}
\subsection{Datasets}
\paragraph{Translated STS-B.}
We use the STS Benchmark development split, comprising 1,500 English sentence pairs annotated on a 0--5 similarity scale \citep{cer2017semeval}. Unique source sentences are translated into ten Indic languages using a Google Translate library in batches of 100. The translated sentences are then mapped back to their original pairs, retaining the English similarity scores. This procedure yields 1,500 pairs per language and 15,000 translated pairs in total. Because all language sets originate from the same source pairs, they do not constitute independent collections of semantic examples.

Table~\ref{tab:languages} summarizes language and script coverage. Retaining the source labels assumes that translation preserves the semantic relationship within each pair. Translation-induced changes to entities, negation, idioms, or specificity may violate this assumption. We therefore distinguish inherited STS-B labels from judgments elicited directly for the translated sentences.

\begin{table}[t]
\centering
\small
\begin{tabular}{@{}lll@{}}
\toprule
Language & Code & Script \\
\midrule
Hindi & hi & Devanagari \\
Marathi & mr & Devanagari \\
Bengali & bn & Bengali \\
Gujarati & gu & Gujarati \\
Kannada & kn & Kannada \\
Malayalam & ml & Malayalam \\
Punjabi & pa & Gurmukhi \\
Tamil & ta & Tamil \\
Telugu & te & Telugu \\
Urdu & ur & Perso-Arabic \\
\bottomrule
\end{tabular}
\caption{Translated STS-B coverage. Each language contains the same 1,500 source pairs, translated into the indicated script.}
\label{tab:languages}
\end{table}

\paragraph{Native Marathi MahaSTS.}
MahaSTS comprises 16,860 human-annotated Marathi sentence pairs with similarity scores on a 0--5 scale \citep{mirashi2025mahasts}. Its published split contains 14,328 training pairs, 1,692 test pairs, and 840 validation pairs. We use the published test split to complement translated STS-B with native-language judgments. MahaSTS and translated STS-B differ in sentence content and annotation; consequently, cross-dataset differences cannot be attributed exclusively to translation.

\subsection{BERT-Based and LLM-Based Models}
Table~\ref{tab:models} lists the thirteen evaluated systems. We use \emph{BERT-based} to denote the broader family of BERT-style encoders and related architectures, including XLM-R and MPNet. The \emph{LLM-based} category comprises embedding models derived from foundation language models. These operational categories describe model lineage and do not imply identical architectures, attention mechanisms, or supervision within a family.

\paragraph{BERT-based encoders.}
This group comprises IndicSBERT, Bhasha-Embed-v0, LaBSE, STS-adapted XLM-R, MuRIL, multilingual e5-base, paraphrase-MPNet-v2, and BGE-M3. We distinguish IndicSBERT and Bhasha-Embed-v0 as Indic-focused sentence embedding models. The remaining encoders include both broadly multilingual systems and MuRIL, an Indic-focused pretrained encoder. This subdivision concerns sentence-level adaptation rather than the presence or absence of Indic pretraining. XLM-R denotes \path{sentence-transformers/stsb-xlm-r-multilingual}, a sentence-level adaptation of XLM-R rather than its unadapted base checkpoint \citep{reimers2019sentence}.

\paragraph{LLM-based embedding models.}
This group comprises Qwen3-Embedding-0.6B, Qwen3-Embedding-4B, EmbeddingGemma-300M, Harrier-oss-v1-270M, and Harrier-oss-v1-0.6B. All are trained for embedding generation, including the compact variants. Because both model families include task-adapted systems with distinct training histories, their comparison assesses empirical STS performance rather than the independent contribution of architecture.

\begin{table*}[t]
\centering
\small
\begin{tabularx}{\textwidth}{@{}l l X@{}}
\toprule
Group & Model & Representation or intended use \\
\midrule
BERT-based: Indic sentence models & IndicSBERT & Multilingual BERT adapted with Indic NLI and STS supervision. \\
 & Bhasha-Embed-v0 & MuRIL-based embeddings targeting Hindi, English, and Romanized Hindi. \\
\midrule
BERT-based: other encoders & LaBSE & Cross-lingually aligned sentence embeddings. \\
 & XLM-R (STS) & Multilingual sentence encoder adapted for STS. \\
 & MuRIL & Indic-focused pretrained contextual encoder. \\
 & Multilingual e5-base & Multilingual contrastive text embeddings. \\
 & paraphrase-MPNet-v2 & Multilingual paraphrase-oriented sentence embeddings. \\
 & BGE-M3 & Multilingual embedding model; only dense cosine scores are considered here. \\
\midrule
LLM-based embedding models & Qwen3-Embedding-0.6B / 4B & Embedding models built on Qwen3 foundation models. \\
 & EmbeddingGemma-300M & Compact multilingual text embedding model. \\
 & Harrier-oss-v1-270M / 0.6B & Decoder-based multilingual embedding models. \\
\bottomrule
\end{tabularx}
\caption{Evaluated models grouped by model family and sentence-level adaptation. BERT-based includes related encoder architectures; LLM-based denotes embeddings derived from foundation language models. Size designations distinguish the evaluated variants.}
\label{tab:models}
\end{table*}

\subsection{Scoring and Aggregation}

\begin{figure}[t]
\centering
\includegraphics[width=0.96\columnwidth]{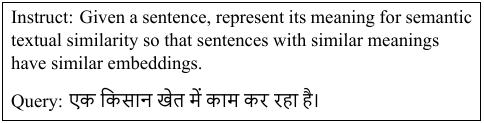}
\caption{Illustrative Qwen3-Embedding-0.6B input with a Hindi query and the English STS instruction. Each sentence in a pair is formatted and encoded independently.}
\label{fig:qwen-input}
\end{figure}

Sentence embeddings are generated independently for each sentence. The XLM-R sentence model is encoded with Sentence Transformers using a batch size of 32, checkpoint-defined mean pooling, and L2 normalization. For Qwen3-Embedding-0.6B, we use Hugging Face Transformers with left padding, truncation at 512 tokens, a batch size of 16, and last-token pooling. Both sentences receive the same English STS instruction in the \texttt{Instruct: \ldots\ Query: \ldots} format (Figure~\ref{fig:qwen-input}). Qwen inference uses FP16 on CUDA and FP32 on CPU; embeddings are converted to FP32 before L2 normalization. Empty sentence pairs and non-numeric or missing labels are excluded in this configuration. For a sentence pair $(x_i,x'_i)$ and embedding function $f$, the predicted similarity is
\begin{equation}
\hat{s}_i=\frac{f(x_i)^\top f(x'_i)}{\lVert f(x_i)\rVert_2\lVert f(x'_i)\rVert_2}.
\end{equation}
We compare predicted similarities with reference scores using Pearson correlation $r$ and, where available, Spearman rank correlation $\rho$. Pearson measures linear association, whereas Spearman measures rank agreement. We report Pearson correlations for all thirteen systems on translated STS-B and Spearman correlations for the two Indic-focused sentence encoders. Both metrics are reported for all systems on MahaSTS. Pearson correlation provides the common metric for comparisons across the complete model set.

For translated STS-B, we summarize performance by the unweighted mean of the ten language-specific Pearson correlations:
\begin{equation}
\bar{r}=\frac{1}{10}\sum_{\ell=1}^{10}r_\ell.
\end{equation}
Each language contributes equally to the macro-average. We aggregate language-specific correlations rather than pooling sentence pairs across languages. Macro-averages are computed from the three-decimal language-level scores and may therefore differ slightly from aggregates based on unrounded predictions.

\paragraph{Interpretation of comparisons.}
Cosine similarity provides the common scoring function across models. Differences reflect the complete training and inference pipelines of the evaluated systems; the comparison does not isolate architecture from these factors.

\section{Results}
\subsection{Translated STS-B Results}
\begin{table*}[t]
\centering
\small
\setlength{\tabcolsep}{2.5pt}
\begin{tabular}{@{}>{\raggedright\arraybackslash}p{4.5cm}*{11}{r}@{}}
\toprule
Model & hi & mr & bn & gu & kn & ml & pa & ta & te & ur & Mean \\
\midrule
\multicolumn{12}{@{}l}{\emph{BERT-based: Indic-focused sentence embedding models}} \\
IndicSBERT \citep{deode2023l3cube} & \textbf{0.853} & \textbf{0.846} & \textbf{0.851} & \textbf{0.842} & \textbf{0.843} & \textbf{0.839} & \textbf{0.838} & \textbf{0.845} & \textbf{0.838} & \textbf{0.838} & \textbf{0.843} \\
Bhasha-Embed-v0 \citep{sukhlecha2024bhasha} & 0.790 & 0.759 & 0.705 & 0.732 & 0.750 & 0.756 & 0.741 & 0.756 & 0.734 & 0.741 & 0.746 \\
\midrule
\multicolumn{12}{@{}l}{\emph{BERT-based: other encoders}} \\
LaBSE \citep{feng2020language} & 0.726 & 0.717 & 0.703 & 0.725 & 0.709 & 0.702 & 0.705 & 0.718 & 0.727 & 0.707 & 0.714 \\
XLM-R \citep{reimers2019sentence} & 0.836 & 0.772 & 0.716 & 0.728 & 0.730 & 0.720 & 0.707 & 0.728 & 0.713 & 0.758 & 0.741 \\
MuRIL \citep{khanuja2021muril} & 0.659 & 0.675 & 0.664 & 0.678 & 0.676 & 0.684 & 0.680 & 0.699 & 0.693 & 0.653 & 0.676 \\
e5-base \citep{wang2024e5} & 0.789 & 0.748 & 0.755 & 0.710 & 0.736 & 0.755 & 0.703 & 0.744 & 0.726 & 0.757 & 0.742 \\
MPNet \citep{song2020mpnet,reimers2020multilingual} & 0.839 & 0.804 & 0.708 & 0.777 & 0.694 & 0.733 & 0.726 & 0.721 & 0.726 & 0.818 & 0.755 \\
BGE-M3 \citep{chen2024bge} & 0.822 & 0.788 & 0.791 & 0.783 & 0.780 & 0.794 & 0.784 & 0.783 & 0.791 & 0.799 & 0.792 \\
\midrule
\multicolumn{12}{@{}l}{\emph{LLM-based embedding models}} \\
Qwen3-0.6B \citep{zhang2025qwen3} & 0.777 & 0.723 & 0.730 & 0.721 & 0.697 & 0.687 & 0.712 & 0.703 & 0.718 & 0.691 & 0.716 \\
Qwen3-4B \citep{zhang2025qwen3} & 0.780 & 0.726 & 0.728 & 0.740 & 0.715 & 0.687 & 0.726 & 0.722 & 0.714 & 0.743 & 0.728 \\
EmbeddingGemma \citep{google2025embeddinggemma} & 0.825 & 0.765 & 0.785 & 0.730 & 0.724 & 0.742 & 0.681 & 0.747 & 0.736 & 0.722 & 0.746 \\
Harrier-270M \citep{microsoft2026harrier} & 0.767 & 0.750 & 0.743 & 0.744 & 0.754 & 0.748 & 0.762 & 0.760 & 0.736 & 0.746 & 0.751 \\
Harrier-0.6B \citep{microsoft2026harrier} & 0.781 & 0.753 & 0.743 & 0.736 & 0.704 & 0.722 & 0.721 & 0.722 & 0.723 & 0.761 & 0.737 \\
\bottomrule
\end{tabular}
\caption{Pearson correlations on translated STS-B. Mean denotes the unweighted macro-average across languages; bold indicates the highest score in each column. Language codes are defined in Table~\ref{tab:languages}.}
\label{tab:translated}
\end{table*}

Table~\ref{tab:translated} reports Pearson correlations across the ten translated language sets. IndicSBERT achieves the highest correlation in every language and a macro-average of 0.843. BGE-M3 ranks second overall at 0.792, with a difference of 0.052 when calculated before rounding the macro-averages. Harrier-270M is the highest-scoring LLM-based model at 0.751, followed by EmbeddingGemma at 0.746. The difference between the highest-scoring models in the two families is 0.092. IndicSBERT's advantage over all evaluated LLM-based systems is consistent across the ten languages (RQ1). IndicSBERT's explicit training on Indic STS datasets may contribute to its higher correlations \citep{deode2023l3cube}.

\paragraph{Variation among BERT-based encoders.}
IndicSBERT's Pearson correlations range from 0.838 to 0.853, while its Spearman correlations range from 0.836 to 0.850 (Table~\ref{tab:indic-ranks}). Its narrow score range indicates consistent performance across the translated sets. Bhasha-Embed-v0 exhibits greater language-level variation, with Pearson correlations ranging from 0.705 to 0.790. Its highest score occurs in Hindi, a stated target language. These observations are compatible with differences in language and task adaptation, but do not establish their causal contributions (RQ2).

Among the other BERT-based encoders, MPNet exceeds BGE-M3 in Hindi, Marathi, and Urdu, despite BGE-M3's higher macro-average. Performance variation within the family thus remains substantial, and the aggregate ranking does not hold uniformly at the language level.

\paragraph{Variation among LLM-based models.}
EmbeddingGemma exceeds Harrier-270M in Hindi, Marathi, and Bengali; Harrier-270M achieves higher scores in six languages, with a tie in Telugu. Consequently, the best-performing LLM-based model depends on the target language.

The Qwen3-4B variant exceeds Qwen3-0.6B in macro-average Pearson correlation (0.728 versus 0.716), although the improvement is not uniform across languages. Conversely, Harrier-270M exceeds Harrier-0.6B on average (0.751 versus 0.737). Larger parameter counts therefore do not yield consistent gains within the evaluated LLM-based families (RQ2). This comparison does not control for differences in training or inference configuration.

\begin{table}[t]
\centering
\small
\begin{tabular}{@{}lrrrr@{}}
\toprule
& \multicolumn{2}{c}{IndicSBERT} & \multicolumn{2}{c}{Bhasha-Embed} \\
\cmidrule(lr){2-3}\cmidrule(l){4-5}
Language & $r$ & $\rho$ & $r$ & $\rho$ \\
\midrule
Hindi & 0.853 & 0.850 & 0.790 & 0.800 \\
Marathi & 0.846 & 0.843 & 0.759 & 0.770 \\
Bengali & 0.851 & 0.850 & 0.705 & 0.728 \\
Gujarati & 0.842 & 0.840 & 0.732 & 0.742 \\
Kannada & 0.843 & 0.841 & 0.750 & 0.761 \\
Malayalam & 0.839 & 0.840 & 0.756 & 0.768 \\
Punjabi & 0.838 & 0.838 & 0.741 & 0.753 \\
Tamil & 0.845 & 0.844 & 0.756 & 0.767 \\
Telugu & 0.838 & 0.836 & 0.734 & 0.744 \\
Urdu & 0.838 & 0.836 & 0.741 & 0.753 \\
\midrule
Mean & 0.843 & 0.842 & 0.746 & 0.759 \\
\bottomrule
\end{tabular}
\caption{Pearson ($r$) and Spearman ($\rho$) correlations for Indic-focused sentence encoders on translated STS-B. Pearson correlations provide the common comparison metric for the complete model set.}
\label{tab:indic-ranks}
\end{table}

\subsection{Native Marathi Results}
Table~\ref{tab:native} reports MahaSTS results alongside Pearson correlations for translated Marathi STS-B. IndicSBERT achieves the highest MahaSTS Pearson correlation (0.901), closely followed by Bhasha-Embed-v0 (0.900). Both obtain a Spearman correlation of 0.890 at the reported precision. The Pearson difference is 0.001; without uncertainty estimates, this numerical ordering does not establish a reliable performance difference.

BGE-M3 and multilingual e5-base achieve Pearson correlations of 0.891 and 0.890, respectively. EmbeddingGemma leads the LLM-based models at 0.871, followed by Harrier-270M at 0.815. The difference between IndicSBERT and EmbeddingGemma is 0.030 on MahaSTS, compared with 0.081 on translated Marathi STS-B. The highest-scoring BERT-based model therefore remains ahead of the highest-scoring LLM-based model on native Marathi, although the gap is smaller (RQ3).

This ordering does not extend to every pair of models across families. MuRIL obtains a lower MahaSTS correlation than all five LLM-based systems, demonstrating substantial overlap in performance between the two families.

\begin{table}[t]
\centering
\small
\setlength{\tabcolsep}{3.5pt}
\begin{tabular}{@{}lrrr@{}}
\toprule
& STS-B & \multicolumn{2}{c}{MahaSTS} \\
\cmidrule(l){3-4}
Model & $r_{\mathrm{mr}}$ & $r$ & $\rho$ \\
\midrule
IndicSBERT & 0.846 & \textbf{0.901} & \textbf{0.890} \\
Bhasha-Embed-v0 & 0.759 & 0.900 & \textbf{0.890} \\
BGE-M3 & 0.788 & 0.891 & 0.881 \\
e5-base & 0.748 & 0.890 & 0.880 \\
MPNet & 0.804 & 0.881 & 0.872 \\
EmbeddingGemma & 0.765 & 0.871 & 0.859 \\
LaBSE & 0.717 & 0.864 & 0.857 \\
XLM-R & 0.772 & 0.841 & 0.832 \\
Harrier-270M & 0.750 & 0.815 & 0.807 \\
Qwen3-4B & 0.726 & 0.814 & 0.810 \\
Harrier-0.6B & 0.753 & 0.801 & 0.798 \\
Qwen3-0.6B & 0.723 & 0.800 & 0.795 \\
MuRIL & 0.675 & 0.757 & 0.762 \\
\bottomrule
\end{tabular}
\caption{BERT-based and LLM-based models on Marathi, sorted by MahaSTS Pearson correlation. Bold marks the highest reported native score per metric, including the Spearman tie. STS-B and MahaSTS contain different sentence pairs and annotations.}
\label{tab:native}
\end{table}

\begin{figure*}[t]
\centering
\includegraphics[width=\textwidth]{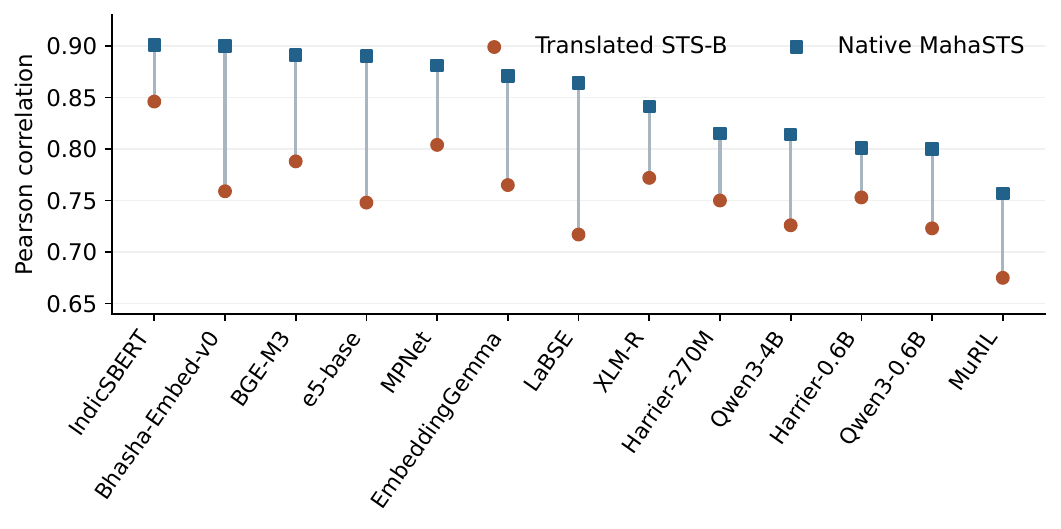}
\caption{Pearson correlations on translated Marathi STS-B and native MahaSTS. Connected markers denote the same model evaluated on different datasets; the differences are descriptive and do not isolate translation effects.}
\label{fig:marathi}
\end{figure*}

\section{Discussion}
\paragraph{Model family and task adaptation.}
The strongest BERT-based encoder achieves higher correlations than the strongest LLM-based embedding model in both evaluation settings. However, the substantial within-family variation indicates that model lineage alone is insufficient to explain STS performance. IndicSBERT's results are consistent with the relevance of Indic STS supervision, while MuRIL's lower scores illustrate the distinction between contextual pretraining and sentence-level adaptation. Since architecture, supervision, and inference are not controlled independently, the evaluation cannot determine their individual contributions.

\paragraph{Sensitivity to benchmark construction.}
All thirteen systems obtain higher Pearson correlations on MahaSTS than on translated Marathi STS-B (Figure~\ref{fig:marathi}). The absolute differences are 0.055 for IndicSBERT, 0.141 for Bhasha-Embed-v0, and 0.106 for EmbeddingGemma. Bhasha-Embed-v0 also moves from below several multilingual alternatives on translated Marathi to nearly matching IndicSBERT on MahaSTS. Thus, both performance gaps and relative rankings depend on the evaluation resource.

These cross-dataset differences admit several explanations, including sentence content, score distributions, annotation procedures, translation quality, and training exposure. They cannot establish that translation systematically underestimates model capability. A matched comparison with direct human judgments of translated pairs would be required to distinguish translation-related effects from other dataset differences.

\paragraph{Implications for model selection.}
IndicSBERT achieves the strongest results under the translated STS-B protocol, while BGE-M3 provides the strongest aggregate alternative among the other BERT-based encoders. EmbeddingGemma is the strongest LLM-based model on native Marathi. These findings support validation on the intended language, domain, and similarity task rather than selection based solely on model family or scale. This task-specific interpretation is consistent with the heterogeneous embedding performance observed in MTEB \citep{muennighoff2023mteb}. Computational efficiency remains outside the scope of the evaluation: parameter count alone does not establish latency, memory use, or deployment cost.

\section{Conclusion}
We evaluated eight BERT-based encoders and five LLM-based embedding models for semantic textual similarity across ten Indic languages. IndicSBERT achieves the highest Pearson correlation in every translated STS-B language set, with a macro-average of 0.843, compared with 0.792 for BGE-M3 and 0.751 for Harrier-270M, the strongest LLM-based model in this setting. On native Marathi MahaSTS, IndicSBERT and Bhasha-Embed-v0 achieve correlations of 0.901 and 0.900, while EmbeddingGemma leads the LLM-based models at 0.871. The results demonstrate that LLM-derived embeddings do not consistently exceed established encoders for Indic STS and that relative performance depends on the evaluation dataset. Further work should expand native-language evaluation and quantify uncertainty and training-data overlap.

\section*{Limitations}
\paragraph{Data and language coverage.}
Translated STS-B inherits English similarity judgments without collecting language-specific annotations. Translation-induced label noise is not quantified, and all ten translated sets share the same source pairs. Native-language evaluation is restricted to Marathi. The findings therefore do not establish performance on other native Indic STS datasets, dialectal or code-mixed text, or specialized domains. MahaSTS construction also uses MahaSBERT-STS scores for candidate selection and as an annotation reference \citep{mirashi2025mahasts}; this may introduce model-dependent selection or annotation effects despite the use of native text.

\paragraph{Training overlap.}
Exposure to STS-B, its translations, MahaSTS, or related data during model training has not been ruled out; evaluation on a development or test split alone does not establish disjointness from pretraining or embedding supervision.

\paragraph{Statistical and causal interpretation.}
We report descriptive correlation estimates and do not infer statistical significance from small numerical differences. Cross-dataset comparisons confound translation with content and annotation, while model comparisons confound architecture with training and inference. The two families also contain unequal numbers of heterogeneous systems. Comparisons between their highest-scoring members do not estimate an intrinsic architectural advantage.

\section*{Ethical Considerations}
Benchmark scores characterize specific datasets and should not be interpreted as measures of a language's complexity or a community's value. Machine translation may alter culturally specific meanings or propagate bias. Redistribution must respect source-data licenses and translation-service terms. No data release is claimed.

\begin{samepage}
\section*{Acknowledgements}
This work was carried out under the mentorship of L3Cube, Pune. We would like to express our gratitude towards our mentor for his continuous support and encouragement. This work is a part of the L3Cube-MahaNLP project \citep{joshi2022mahanlp}.

\end{samepage}

\bibliography{main}

\end{document}